\documentclass{article}

\usepackage[preprint]{neurips_2026}

\usepackage[utf8]{inputenc} 
\usepackage[T1]{fontenc}    
\usepackage{hyperref}       
\usepackage{url}            
\usepackage{booktabs}       
\usepackage{amsfonts}       
\usepackage{amsmath}
\usepackage{amssymb}
\usepackage{nicefrac}       
\usepackage{microtype}      
\usepackage{xcolor}         
\usepackage{enumitem}
\usepackage{verbatim}

\usepackage{listings}
\usepackage{algorithm}
\usepackage[noend]{algorithmic}
\usepackage{graphicx}
\usepackage{multirow}
\usepackage{xcolor}
\usepackage{colortbl}
\usepackage{tcolorbox}
\usepackage{listings}
\usepackage{wrapfig}
\usepackage{pifont}
\lstdefinestyle{prompt}{
  basicstyle=\ttfamily\footnotesize,
  frame=single,
  framesep=6pt,
  rulecolor=\color{black!50},
  backgroundcolor=\color{black!2},
  breaklines=true,
  breakatwhitespace=true,
  columns=fullflexible,
  keepspaces=true,
  showstringspaces=false,
  upquote=true,
  xleftmargin=4pt,
  xrightmargin=4pt,
  aboveskip=6pt,
  belowskip=6pt,
}

\title{MIND-Skill: Quality-Guaranteed Skill Generation via Multi-Agent Induction and Deduction}

\author{
\begin{tabular}{c}
Yixuan Li$^{1}$\thanks{Equal contribution.} \quad
Mingshu Cai$^{2}$\footnotemark[1] \quad
Ziyang Xiao$^{3}$ \quad
Wanyuan Wang$^{4}$ \\[3pt]
Yanchen Deng$^{1}$\thanks{Correspondence to: \texttt{ycdeng@ntu.edu.sg}} \quad
Bo An$^{1}$ \\[6pt]
{\normalfont\small
$^1$Nanyang Technological University \quad
$^2$Waseda University} \\[-1pt]
{\normalfont\small
$^3$Zhejiang University \quad
$^4$Southeast University}
\end{tabular}
}

\definecolor{hlgreen}{RGB}{220,245,220}
\definecolor{hlgreen}{RGB}{220,245,220}
\definecolor{boxborder}{RGB}{180,180,180}
\definecolor{boxbg}{RGB}{250,250,250}
\definecolor{hlblue}{RGB}{220,235,252}
\definecolor{hlorange}{RGB}{255,230,200}
\definecolor{contextbg}{RGB}{243,243,243}

\definecolor{promptframe}{RGB}{70,130,180}
\definecolor{prompttitle}{RGB}{210,230,245}

\newtcolorbox{promptbox}[1]{
  colback=white,
  colframe=promptframe,
  colbacktitle=prompttitle,
  coltitle=black,
  title={\small\textbf{#1}},
  fonttitle=\sffamily,
  boxrule=0.6pt,
  arc=1.5pt,
  toptitle=4pt, bottomtitle=4pt,
  left=5pt, right=5pt, top=4pt, bottom=4pt,
  boxsep=2pt,
  middle=3pt
}

\begin{document}

\maketitle

\begin{abstract}
Large language model (LLM) powered AI agents have emerged as a promising paradigm for autonomous problem-solving, yet they continue to struggle with complex, multi-step real-world tasks that demand domain-specific procedural knowledge. Reusable agent skills, which encapsulate successful problem-solving strategies, offer a natural remedy by enabling agents to build on prior experience. However, curating such skills has largely remained a manual endeavor, requiring human experts to distill rich domain knowledge into actionable guidelines.  In this work, we present \textbf{M}ulti-agent \textbf{IN}duction and \textbf{D}eduction for \textbf{Skill}s (\textbf{MIND-Skill}), a framework that automatically induces generalizable skills from successful trajectories with robust quality guarantees. MIND-Skill consists of an induction agent which is tasked to abstract reusable skills from successful trajectories, and a deduction agent which aims to reconstruct trajectories by following the induced skills. To guarantee the quality of the generated skills, we introduce a reconstruction loss that compares input and reconstructed trajectories, an outcome loss that enforces the correctness of the reconstructed trajectories, and a rubric loss that assesses the documentation quality and regularizes the abstraction level of the generated skills according to predefined criteria. These textual losses are jointly optimized with TextGrad, and the resulting skills are evaluated on held-out tasks unseen during optimization. Experiments on AppWorld and BFCL-v3 show that MIND-Skill consistently outperforms concurrent skill generation methods.

\end{abstract}

\section{Introduction}
\label{sec:intro}
Large language models (LLMs) have demonstrated exceptional performance on various challenging reasoning tasks, including theorem proving~\citep{yang2023leandojo,hubert2026olympiad}, code generation~\citep{lyu2025reloc,wang2025plansearch}, and scientific discovery~\citep{novikov2025alphaevolve}. Equipped with tools, memory, and harness scaffolding, LLM-powered AI agents~\citep{openclaw2026,hermesagent2026,claudecode2025,claudemanagedagents2026} have emerged as a promising paradigm for autonomous problem-solving in many open-ended scenarios. While LLMs inherit extensive declarative knowledge from pretraining, AI agents still struggle with complex, long-horizon tasks that demand domain-specific \textit{procedural knowledge}, such as using APIs, making multi-step tool calls, and adapting actions based on workflow feedback~\citep{trivedi2024appworld, patil2025bfcl}.

Agent skills~\citep{anthropic2025skills}, which encapsulate successful problem-solving strategies and standard operating procedures into bundles of Markdown documents and related scripts, offer an elegant solution by enabling agents to build on prior domain experience~\citep{tagkopoulos2025skillflow,li2026organizing}. However, curating high-quality skills has largely remained a manual endeavor, requiring extensive human expertise to distill rich domain knowledge into actionable guidelines~\citep{li2026skillsbench}. Recent research efforts have attempted to generate skills automatically from different sources of knowledge. Zero-shot techniques~\citep{anthropic2025skillcreator} turn task descriptions or user prompts directly into skills by eliciting the prior knowledge of LLMs, though their effectiveness remains limited~\citep{li2026skillsbench}. Trajectory-distillation methods~\citep{ni2026trace2skill,wang2026skillx,tu2026d2skill} derive reusable skills for novel tasks by abstracting existing execution traces into generalizable procedures, typically in an offline fashion. Lastly, lifelong evolving methods~\citep{hermesagent2026,xia2026skillrl,wang2025sage,alzubi2026evoskill} continuously crystallize and refine skills according to agents' accumulated experiences and memory.

Unfortunately, a key limitation of existing skill generation methods is the lack of quality guarantees. \textbf{First}, many techniques directly generate skills from task specifications, trajectories or experiences without a principled closed-loop pipeline that explicitly validates, corrects, and refines the skills based on execution outcomes. \textbf{Second}, the documentation quality of the generated skills is largely overlooked. Skills are intended to be reusable, portable artifacts that can be shared across agents, models and even human practitioners, yet current methods rarely evaluate whether the produced documents adhere to established standards of technical writing, e.g., logical flow and troubleshooting guidance. \textbf{Third}, for trajectory-distillation methods, the faithfulness of the abstraction process is never verified. Distilling execution traces into reusable skills necessarily involves lossy compression, which potentially leads to over-generalization. Yet there is no established mechanism to guarantee the generated skills faithfully preserve the essential aspects of their source trajectories, such as edge-case handling and prerequisite checks.

In light of this, we propose \textbf{M}ulti-agent \textbf{IN}duction and \textbf{D}eduction for \textbf{Skill}s (\textbf{MIND-Skill}), a novel framework that  synthesizes generalizable skills with quality guarantees from agents' successful trajectories. Unlike existing trajectory-distillation methods that synthesize skills solely from traces, MIND-Skill features an \textit{induction agent} that derives skills from input trajectories, and a \textit{deduction agent} that reconstructs the input trajectories by actively following the generated skills. The faithfulness of the generated skills is therefore enforced by optimizing a \textit{reconstruction loss} that measures the discrepancy between the input trajectories and the reconstructed ones. In addition, we introduce an \textit{outcome loss} that enforces the correctness of the reconstructed trajectories, and a \textit{rubric loss} that assesses documentation quality and regularizes the abstraction level of the generated skills according to predefined criteria. These textual losses are jointly optimized with TextGrad~\citep{yuksekgonul2025textgrad} to produce high-quality skills. Specifically, we make the following contributions:

\begin{itemize}[leftmargin=*]
\item We propose MIND-Skill, a multi-agent
induction and deduction framework that automatically synthesizes generalizable skills from successful trajectories. To ensure that the generated skills carry all critical procedural knowledge, we keep the deduction agent frozen so that it receives no guidance beyond the induced skill when reconstructing trajectories.

\item To guarantee the quality of induced skills, we propose three textual losses and jointly optimize them with TextGrad: a reconstruction loss that measures the discrepancy between the input and reconstructed trajectories, an outcome loss that enforces the execution correctness, and a rubric loss that assesses documentation quality and regularizes the abstraction level of the generated skills.

\item We evaluate MIND-Skill on AppWorld~\citep{trivedi2024appworld} and BFCL-v3~\citep{patil2025bfcl}, and show that the induced skills improve agent performance on both source tasks and held-out tasks unseen during generation.


\end{itemize}



\section{Related Work}
\label{sec:related}
\subsection{Agent Skills}
\label{sec:rw_skill}
Agent skills encapsulate reusable procedural knowledge into structured documents that can be shared across agents, models, and even human practitioners~\citep{anthropic2025skills}. Recent surveys systematize the skill lifecycle and distinguish skills from generic tool use by their procedural, reusable nature~\citep{jiang2026sok,xu2026agentskills}. \citet{li2026organizing} demonstrate that single agents augmented with in-depth skills can match the performance of multi-agent frameworks. That said, the mere presence of skills does not guarantee improved performance. SkillsBench~\citep{li2026skillsbench} reveals that zero-shot-generated skills provide no benefit on average, whereas agents equipped with curated, human-authored skills consistently outperform the no-skill baseline. SWE-Skills-Bench~\citep{han2026sweskills} further demonstrates that low-quality skills can significantly degrade agent performance rather than improve it. Our work directly tackles this gap by coupling skill induction with deduction-based verification, providing closed-loop quality guarantees for generated skills.
\subsection{Skill Generation}
\label{sec:rw_generation}
\paragraph{Zero-shot generation.}
Zero-shot methods produce skills directly from task descriptions or user prompts by eliciting the parametric knowledge of LLMs~\citep{anthropic2025skillcreator}, without leveraging any execution experience. While lightweight, these methods are fundamentally limited by the absence of execution experience and therefore cannot capture domain-specific procedural knowledge that only emerges through step-by-step interaction with the environment~\citep{li2026skillsbench}.

\paragraph{Trajectory distillation.}
Trajectory-distillation methods abstract execution traces into reusable agent skills.
WebXSkill~\citep{wang2026webxskill} extracts reusable action subsequences from synthetic agent trajectories and abstracts them into parameterized skills that pair executable action programs with step-level natural language guidance.
Trace2Skill~\citep{ni2026trace2skill} dispatches parallel sub-agents to extract trajectory lessons and then hierarchically consolidates them into a skill directory.
SkillX~\citep{wang2026skillx} extracts a three-level skill hierarchy from rollout trajectories and refines it via merging and filtering.
D2Skill~\citep{tu2026d2skill} reflects on execution trajectories to generate skills at both task and step granularities.
While these methods differ in abstraction strategies, they share two common limitations: the faithfulness of the abstraction process is never explicitly verified, and the documentation quality of the generated skills is largely uncontrolled. MIND-Skill addresses both gaps by requiring a frozen deduction agent to reconstruct the source trajectories from the generated skill alone, which provides an explicit faithfulness signal, and by introducing a rubric loss that enforces documentation standards and regularizes the abstraction level.



\paragraph{Lifelong evolving methods.}
Lifelong methods continuously generate and refine skills from accumulated experience. SAGE~\citep{wang2025sage} and
SkillRL~\citep{xia2026skillrl} apply reinforcement learning to improve skills from environment feedback, but produce skills tightly coupled to a specific policy.
EvoSkill~\citep{alzubi2026evoskill} proposes new skills from execution failures and retains them via Pareto-frontier selection. CoEvoSkills~\citep{zhang2026coevoskills} co-evolves a skill generator with a surrogate verifier that provides feedback without ground-truth tests. ACE~\citep{zhang2026ace} accumulates strategies into an evolving playbook through generation-reflection-curation loops. Although these methods leverage environment feedback, the resulting signal is confounded by the agent's own reasoning ability: a capable agent may succeed despite a poor skill, while a weaker agent may fail despite adequate guidance. MIND-Skill disentangles these factors through controlled reconstruction, isolating skill quality as the sole objective and enabling principled optimization via TextGrad.

\begin{figure*}[t]
\centering
\includegraphics[width=\textwidth]{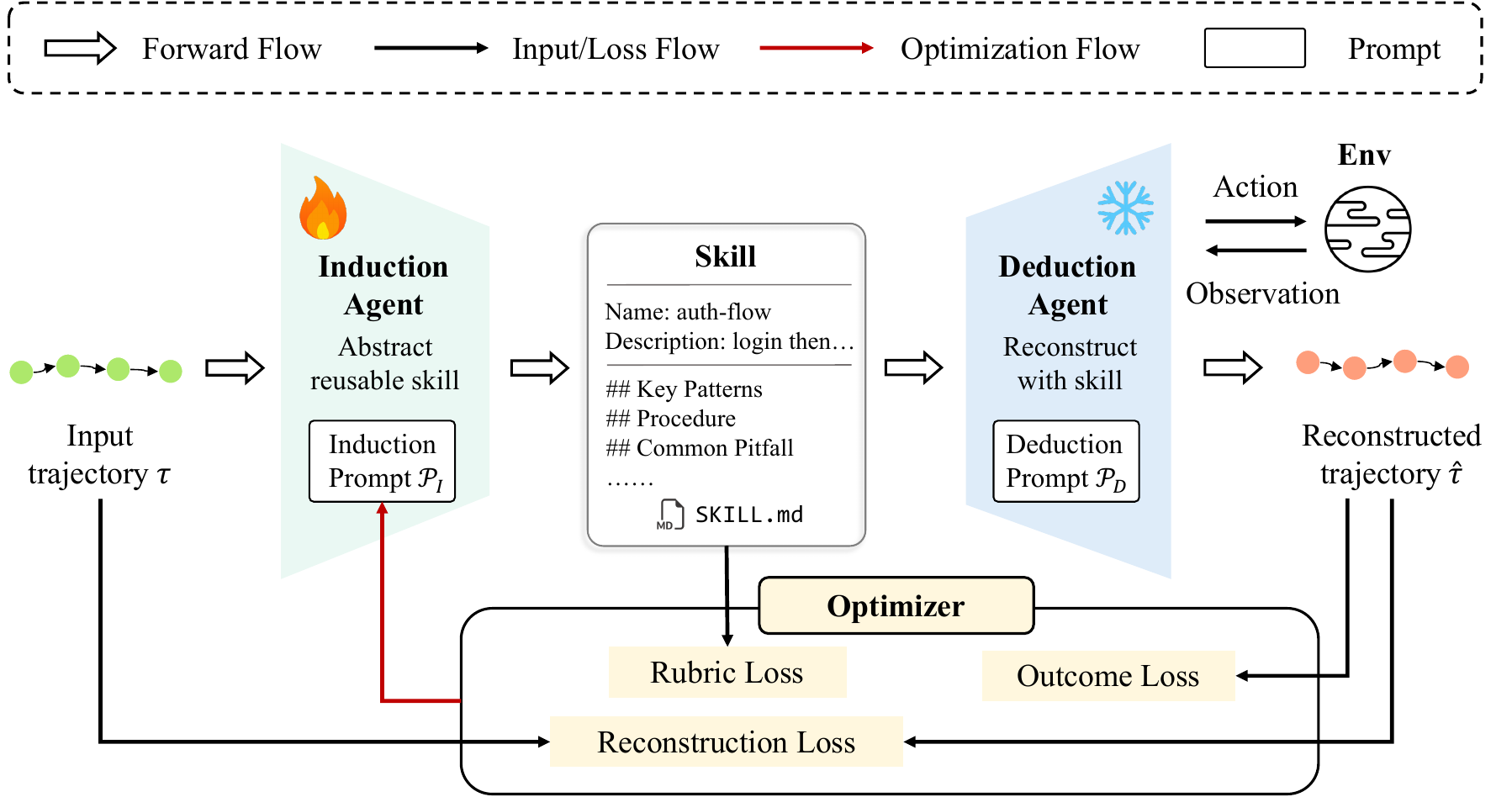}
\caption{\textbf{Overview of MIND-Skill.} The \textbf{induction agent} $\mathcal{A}_I$ (with optimizable prompt $\mathcal{P}_I$) abstracts a successful trajectory $\tau$ into a structured skill document. The \textbf{deduction agent} $\mathcal{A}_D$ (with frozen prompt $\mathcal{P}_D$) then attempts to reconstruct the trajectory by following only the induced skill and the task specification in a live environment. Three textual losses assess the quality of the generated skill: the \textbf{reconstruction loss} measures procedural alignment between $\tau$ and $\hat{\tau}$, the \textbf{outcome loss} evaluates the outcome correctness of $\hat{\tau}$ against the environment, and the \textbf{rubric loss} assesses the documentation quality and regularizes the abstraction level of the skill itself. The text-based \textbf{optimizer} aggregates their textual feedback to update the induction prompt $\mathcal{P}_I$ via TextGrad. Task specification $t$ is omitted from the figure for visual clarity.}
\label{fig:overview}
\end{figure*}

\section{MIND-Skill}
\label{sec:method}
Successful trajectories contain valuable procedural knowledge, yet mining high-quality, generalizable agent skills from them is inherently challenging since they often entangle transferable strategies with instance-level details. MIND-Skill addresses this issue with a novel multi-agent induction and deduction framework. Specifically, the \textit{induction agent} $\mathcal{A}_I$, with an optimizable prompt $\mathcal{P}_I$, is tasked with deriving a skill $s$ from an input (successful) trajectory $\tau$ and the task specification $t$, while the \textit{deduction agent} $\mathcal{A}_D$ attempts to reconstruct $\tau$ solely according to $t$ and $s$. To ensure $s$ preserves all critical procedural knowledge, we keep the deduction agent's prompt $\mathcal{P}_D$ frozen so that it receives no guidance beyond the induced skill during reconstruction and optimization.

For each input pair $(t,\tau)$, we optimize the induction prompt $\mathcal{P}_I$ with respect to three textual loss functions: a \textit{reconstruction loss} $\mathcal{L}_\text{recon}$ that measures procedural alignment between the original and reconstructed trajectories, an \textit{outcome loss} $\mathcal{L}_\text{outcome}$ that enforces the correctness of the reconstructed trajectory, and a \textit{rubric loss} $\mathcal{L}_\text{rubric}$ that assesses documentation quality and regularizes the abstraction level of the skill. For each input task $t$ and trajectory $\tau$, we perform \textbf{lexicographic minimization} where  $\mathcal{L}_\text{outcome}$ is the primary objective, with $\mathcal{L}_\text{recon}$ and $\mathcal{L}_\text{rubric}$ as successive tiebreakers. Formally, 
\begin{equation}
\begin{aligned}
\mathcal{P}_I^* = \mathop{\arg\min}_{\mathcal{P}_I}\;\bigl(
  &\mathcal{L}_\text{outcome}(\hat{\tau},t),\,\mathcal{L}_\text{recon}(\tau,\hat{\tau},t),\,
   \mathcal{L}_\text{rubric}(s,t)\bigr), \\
\text{s.t.}\quad
  s &= \mathcal{A}_I(t,\tau;\mathcal{P}_I), \quad
  \hat{\tau} = \mathcal{A}_D(t,s;\mathcal{P}_D),
\end{aligned}
\end{equation}
and the final skill is given by $s^*=\mathcal{A}_I(t,\tau;\mathcal{P}_I^*)$. 

An overview of MIND-Skill is illustrated in Figure~\ref{fig:overview}. In the following, we describe the induction agent (\S\ref{sec:induction}), the deduction agent (\S\ref{sec:deduction}), the loss functions (\S\ref{sec:loss}), and the optimization procedure (\S\ref{sec:optim}).

\subsection{Induction Agent}
\label{sec:induction}
The induction agent $\mathcal{A}_I$ abstracts a
successful trajectory into a reusable agent skill. The core challenge is controlling the level of abstraction. An over-specific skill that retains instance-level details (e.g., concrete API field paths or entity identifiers) may ease reconstruction of the source task but fails to generalize across task variations. Conversely, an over-abstract skill that merely states high-level intent provides no procedural guidance beyond the task specification itself. The induction agent must therefore identify and preserve only the \textit{non-obvious procedural structure} that occupies the middle ground between these two failure modes.
 
Formally, the induction agent $\mathcal{A}_I$ is parameterized by a system prompt $\mathcal{P}_I$, which is the sole variable optimized during refinement (\S\ref{sec:optim}). Given a task specification $t$ and a successful ReAct~\citep{yao2023react} trajectory $\tau = \{(\text{thought}_m, \text{code}_m, \text{observation}_m)\}_{m=1}^{|\tau|}$, it produces a structured skill document $s = \mathcal{A}_I(t, \tau; \mathcal{P}_I)$. To enforce the desired abstraction level, $\mathcal{P}_I$ encodes a taxonomy that partitions candidate claims into three categories: (1)~\textit{procedural conventions} that generalize across tasks but are non-trivial to infer without execution experience (e.g., paginate until the response is empty), (2)~\textit{instruction-inferable} knowledge derivable from $t$ alone (e.g., an aggregation task implies a counting or grouping operation), and (3)~\textit{ground-truth leakage} that is only knowable from $\tau$ (e.g., concrete response schemas, library choices, or hard-coded thresholds). The prompt $\mathcal{P}_I$ directs $\mathcal{A}_I$ to retain only non-obvious patterns from category~(1), while explicitly suppressing (2) and~(3). This taxonomy serves as the primary inductive bias that TextGrad refines across optimization iterations.

\subsection{Deduction Agent}
\label{sec:deduction}

The deduction agent $\mathcal{A}_D$ reconstructs the trajectory from the induced skill alone. Its prompt $\mathcal{P}_D$ is frozen throughout optimization and receives no access to the source trajectory $\tau$, ensuring that any improvement in reconstruction quality is solely attributable to the skill $s$. Concretely, given the skill $s$ and the task specification $t$, the deduction agent executes a multi-step ReAct loop in a live environment to produce a reconstructed trajectory $\hat{\tau} = \mathcal{A}_D(t, s; \mathcal{P}_D)$. At each step, the agent reasons about the next action, executes code, and observes the environment response. The skill is injected into the agent's prompt as a procedural playbook, serving as the only source of strategic guidance.

\subsection{Textual Loss Functions}
\label{sec:loss}

Existing methods typically refine skills by diagnosing
errors from failed trajectories against reference solutions and incorporating the lessons back into skills. However, a capable agent may compensate for skill deficiencies through its own reasoning, masking gaps that should be fixed, while a weak agent may fail despite adequate guidance, producing misleading negative signals. In either case, task performance
becomes an unreliable proxy for skill quality. Our
reconstruction-based design provides a controlled
alternative: rather than diagnosing failures post-hoc, we directly test whether the skill alone can reproduce the procedural structure of the reference trajectory. Because the deduction agent is frozen and receives no strategic guidance beyond the induced skill, divergences between $\hat{\tau}$ and $\tau$ can be directly attributed to deficiencies in $s$, yielding a clean signal for optimizing the induction agent.  We formalize this through three complementary losses:

\paragraph{Reconstruction loss.}
\label{sec:recon}
The reconstruction loss evaluates whether the induced skill $s$ preserves the essential problem-solving strategy of the source trajectory $\tau$. An LLM judge $\mathcal{A}_J$ takes the reconstructed trajectory $\hat{\tau}$, the source trajectory $\tau$, and the task specification $t$ as inputs, then produces a scalar loss value along with textual feedback:
\begin{equation}
\label{eq:recon}
\mathcal{L}_{\text{recon}}(\tau, \hat{\tau},t)
= \bigl(\, \ell_{\text{recon}},\; f_{\text{recon}} \,\bigr)
= \mathcal{A}_J(\tau, \hat{\tau},t;\mathcal{P}_{\text{recon}}),
\end{equation}
where $\ell_{\text{recon}}\in[0,10]$ measures trajectory discrepancy, $f_{\text{recon}}$ is a natural-language critique identifying specific mismatches, and $\mathcal{P}_{\text{recon}}$ is the system prompt instructing $\mathcal{A}_J$ to compare these two trajectories. Crucially, the judge evaluates tactic-level equivalence rather than step-level similarity: two trajectories that use different API endpoints, loop constructs, or intermediate variables are considered aligned as long as they implement the same procedural logic (e.g., the same retrieval-then-aggregation pattern, the same pagination strategy, or the same prerequisite checking order). 
\paragraph{Outcome loss.}
\label{sec:exec}
The outcome loss provides the only ground-truth signal in our framework by executing the reconstructed trajectory $\hat{\tau}$ in a live environment:
\begin{equation}
\label{eq:outcome}
\mathcal{L}_{\text{outcome}}(\hat{\tau}, t)
= \bigl(\, \ell_{\text{outcome}},\; f_{\text{outcome}} \,\bigr)
= \text{EnvExec}(\hat{\tau},t),
\end{equation}
where $\ell_{\text{outcome}}\in[0,1]$ measures the degree of task failure and
$f_{\text{outcome}}$ captures environment feedback such as
error messages and execution traces. Unlike the
reconstruction loss, which relies on LLM judgment to assess faithfulness of the skill induction process, this signal is grounded in actual task execution and provides a complementary anchor from the perspective of outcome correctness.

\paragraph{Rubric loss.}
\label{sec:rubric}
The rubric loss evaluates the skill document $s$ along two axes. The first is \textit{documentation quality}: whether the skill adheres to established standards of technical writing, such as logical flow, troubleshooting guidance, and completeness, ensuring that it serves as a reusable, portable artifact. The second is \textit{level of abstraction}: the reconstruction and outcome losses optimize for faithful and correct reproduction of the source trajectory, but they cannot distinguish a genuinely transferable skill from one that simply memorizes implementation details. The rubric loss addresses this by detecting statements in the
skill that are tied to the specific implementation of the source trajectory rather than to transferable procedural patterns. Formally,
\begin{equation}
\label{eq:rubric}
\mathcal{L}_\text{rubric}(s,t)=(\ell_{\text{rubric}},\; f_{\text{rubric}})
= \mathcal{A}_J(s,t;\mathcal{P}_{\text{rubric}}),
\end{equation}
where $\ell_{\text{rubric}}\in[0,10]$ denotes rubric violation degree across five dimensions: whether the skill avoids implementation details tied to the source trajectory (\textit{ground-truth independence}), whether it provides sufficient procedural guidance to act on (\textit{actionability}), whether it applies to structurally similar tasks beyond the source (\textit{transferability}), whether all key procedural stages are covered (\textit{completeness}), and whether it is free of redundant boilerplate (\textit{conciseness}). $f_{\text{rubric}}$ provides textual feedback identifying specific issues along these dimensions. The rubric loss serves as a regularizer on abstraction level: without it, the induction agent can inflate reconstruction and execution performance by injecting instance-specific details into the skill, making the skill fail to generalize to novel tasks.
\renewcommand{\algorithmicrequire}{\textbf{Input:}}
\renewcommand{\algorithmicensure}{\textbf{Output:}}
\algsetup{indent=1.8em}
\begin{algorithm}[t]
\caption{Multi-agent Induction and Deduction for Skills (MIND-Skill)}
\label{alg:MIND-Skill}
\begin{algorithmic}[1]
\REQUIRE Task specification $t$, successful trajectory $\tau$, initial induction prompt $\mathcal{P}_I^{(0)}$, frozen deduction prompt $\mathcal{P}_D$, maximum number of iterations $Q$
\ENSURE Optimized skill $s^*$

\STATE $s^* \leftarrow \texttt{nil},\quad \ell_{\text{recon}}^* \leftarrow \infty, \quad\ell_{\text{outcome}}^* \leftarrow \infty, \quad\ell_{\text{rubric}}^* \leftarrow \infty$

\FOR{$q = 0, 1, \ldots, Q-1$}

    \STATE \textcolor{gray}{\textit{\# Induction: distill trajectory into skill}}
    \STATE $s \leftarrow \mathcal{A}_I(t, \tau;\; \mathcal{P}_I^{(q)})$

    \STATE \textcolor{gray}{\textit{\# Deduction: reconstruct trajectory from skill and task specification}}
    \STATE $\hat{\tau} \leftarrow \mathcal{A}_D(t, s;\; \mathcal{P}_D)$

    \STATE \textcolor{gray}{\textit{\# Compute textual losses (each returns a loss value and textual feedback)}}
    \STATE $(\ell_{\text{recon}},\; f_{\text{recon}}) \leftarrow \mathcal{L}_{\text{recon}}(\tau,\; \hat{\tau},\;t)$
    \STATE $(\ell_{\text{outcome}},\; f_{\text{outcome}}) \leftarrow \mathcal{L}_{\text{outcome}}(\hat{\tau},\; t)$
    \STATE $(\ell_{\text{rubric}},\; f_{\text{rubric}}) \leftarrow \mathcal{L}_{\text{rubric}}(s,t)$

    \STATE \textcolor{gray}{\textit{\# Track the best skill with lexicographic comparison}}
    \IF{$(\ell_{\text{outcome}},\ell_{\text{recon}},\ell_{\text{rubric}}) <_{\textbf{lex}} (\ell_{\text{outcome}}^*,\ell_{\text{recon}}^*,\ell_{\text{rubric}}^*)$}
        \STATE $s^* \leftarrow s,\quad \ell_{\text{recon}}^* \leftarrow \ell_{\text{recon}}, \quad\ell_{\text{outcome}}^* \leftarrow \ell_{\text{outcome}}, \quad\ell_{\text{rubric}}^* \leftarrow \ell_{\text{rubric}}$
    \ENDIF

    \STATE \textcolor{gray}{\textit{\# TextGrad: compute textual gradient and update induction prompt}}
    \STATE  $g \leftarrow \text{GradientLLM}\!\left(\mathcal{P}_I^{(q)},t,\; s,\;\hat{\tau},\; f_{\text{recon}},\; f_{\text{outcome}},\; f_{\text{rubric}}\right)$
    \STATE $\mathcal{P}_I^{(q+1)} \leftarrow \text{OptimizerLLM}\!\left(\mathcal{P}_I^{(q)},\; g\right)$

\ENDFOR

\RETURN $s^*$
\end{algorithmic}
\end{algorithm}

\subsection{Closed-Loop Optimization}
\label{sec:optim}
We optimize the induction prompt $\mathcal{P}_I$ to improve the induced skill $s$ through iterative textual gradient descent following TextGrad~\citep{yuksekgonul2025textgrad}. A key design choice is that the gradient LLM observes the reconstructed trajectory $\hat{\tau}$ but not the source trajectory $\tau$: information about $\tau$ reaches the gradient only indirectly through the reconstruction feedback $f_{\text{recon}}$. This prevents the optimizer from proposing superficial fixes that copy implementation details from $\tau$ into the prompt, and together with the rubric loss forms a dual safeguard against ground-truth leakage in the optimization process. Concretely, a gradient LLM consumes the current prompt $\mathcal{P}_I^{(q)}$, the task specification $t$, the induced skill $s$, the reconstructed trajectory $\hat{\tau}$, and the textual feedback from all three losses, and synthesizes a natural-language gradient $g$ that diagnoses failure patterns and proposes revisions. Then an optimizer LLM applies $g$ to produce an updated prompt $\mathcal{P}_I^{(q+1)}$ for the induction agent.

The full procedure is summarized in Algorithm~\ref{alg:MIND-Skill}. For each input pair $(t, \tau)$, we iterate for up to $Q$ iterations: the current prompt $\mathcal{P}_I^{(q)}$ instructs the induction agent to derive a skill $s$ from $\tau$ (line~4), the frozen deduction agent reconstructs the trajectory in a live environment (line~6), and the three losses evaluate the skill and trajectories (lines~8--10). We track the best skill $s^*$ across iterations by lexicographic comparison (lines~12--13) to ensure the anytime property~\citep{zilberstein1996using}. Finally, the textual feedback drives prompt update for the induction agent via TextGrad (lines~15--16).

\section{Experiments}
\label{sec:experiments}


\paragraph{Benchmarks.} 
We evaluate on two complex, long-horizon benchmarks. \textbf{AppWorld}~\citep{trivedi2024appworld} is an interactive coding agent benchmark comprising 9 daily-life apps and 457 APIs. Tasks are officially partitioned into train, test-normal, and test-challenge splits; we extract skills from the 90 training tasks and evaluate on both test splits (168 normal, 417 challenge). We report Task Goal Completion (TGC), the fraction of tasks where all unit tests pass, and Scenario Goal Completion (SGC), which requires all task variations within a scenario to pass. \textbf{BFCL-v3}~\citep{patil2025bfcl} is a multi-turn function-calling benchmark. We use the base multi-turn category (200 instances), randomly split into 50 training and 150 test instances.

\paragraph{Baselines.} We consider the following baselines for comparison:  \textbf{(i) ReAct}~\citep{yao2023react} uses a task prompt with a single demonstration example. For AppWorld, we follow the official ReAct implementation; for BFCL, we use the benchmark's native function-calling mode; \textbf{ (ii) In-Context Learning (ICL)}~\citep{agarwal2024ICL} provides the model with diverse task demonstrations in the input prompt, allowing it to infer task format and desired output; \textbf{(iii) Skill-extract} uses the same induction agent as MIND-Skill to extract a skill from the source trajectory in a single pass without any iterative optimization. This serves as an ablation that isolates the contribution of our closed-loop optimization;  \textbf{(iv) ACE}~\citep{zhang2026ace} is a recent lifelong evolving method that accumulates strategies into a monolithic playbook through generation-reflection-curation loops. We use the official codebase in the offline adaptation mode with ground-truth solutions available during training; implementation details are provided in Appendix~\ref{app:ace}; \textbf{(v) Trace2Skill}~\citep{ni2026trace2skill} is a concurrent method that converts execution traces into structured skills through parallel analysis and hierarchical merge. We use its official codebase; implementation details are provided in
Appendix~\ref{app:trace2skill}.
\paragraph{MIND-Skill implementation.} For each training task, we roll out a successful trajectory with a frontier model, which serves as input to for MIND-Skill. We use the same base model for induction agent, gradient LLM, and optimizer LLM. The maximum number of iterations is set to $Q=8$. For each test task, we prompt the LLM to retrieve $K=3$ skills from the generated skills, and inject them into the LLM's context before executing ReAct loop. Further details are provided in Appendix~\ref{app:mind_skill_impl}.

\begin{table*}[t]
\centering
\caption{Main results on AppWorld and BFCL-v3. All methods use Qwen3.5-122B-A10B for inference. \textbf{Bold} indicates the best and \underline{underline} indicates the second-best result per group.}
\label{tab:main}
\vspace{2mm}
\setlength{\tabcolsep}{6pt}
\begin{tabular}{l cc cc c c}
\toprule
& \multicolumn{2}{c}{\textbf{AppWorld-Normal}} & \multicolumn{2}{c}{\textbf{AppWorld-Challenge}} & \textbf{BFCL-v3} & \\
\cmidrule(lr){2-3} \cmidrule(lr){4-5} \cmidrule(lr){6-6}
\multirow{-2}{*}{\textbf{Method}} & TGC$\uparrow$ & SGC$\uparrow$ & TGC$\uparrow$ & SGC$\uparrow$ & ACC$\uparrow$ & \multirow{-2}{*}{\textbf{Average}} \\
\midrule
\rowcolor[gray]{0.92} \multicolumn{7}{c}{\textit{No skill augmentation}} \\
\addlinespace[2pt]
\textbf{ReAct}          & 55.4 & 35.7 & 36.7 & 23.0 & 63.3 & 42.8 \\
\textbf{ICL}            & 58.9 & 39.3 & 42.7 & 26.6 & 64.7 & 46.4 \\
\midrule
\rowcolor[gray]{0.92} \multicolumn{7}{c}{\textit{Skills generated by Qwen3.5-122B-A10B}} \\
\addlinespace[2pt]
\textbf{Skill-extract}  & 63.7 & 46.4 & 45.1 & 30.9 & 68.7 & 51.0 \\
\textbf{ACE}            & 65.5 & \textbf{55.4} & \underline{51.1} & \underline{34.5} & \underline{74.0} & \underline{56.1} \\
\textbf{Trace2Skill}    & \underline{67.3} & \textbf{55.4} & 46.8 & 33.1 & 72.7 & 55.1 \\
\textbf{MIND-Skill (ours)}   & \textbf{71.4} & \textbf{55.4} & \textbf{51.8} & \textbf{39.6} & \textbf{77.3} & \textbf{59.1} \\
\midrule
\rowcolor[gray]{0.92} \multicolumn{7}{c}{\textit{Skills generated by GPT-5.4}} \\
\addlinespace[2pt]
\textbf{Skill-extract}  & 63.7 & 44.6 & 47.7 & 33.1 & 69.3 & 51.7 \\
\textbf{ACE}            & 69.0 & \textbf{58.9} & 47.2 & 33.8 & \underline{72.7} & 56.3 \\
\textbf{Trace2Skill}    & \underline{70.2} & 51.8 & \textbf{52.8} & \underline{36.7} & 71.3 & \underline{56.6} \\
\textbf{MIND-Skill (ours)}   & \textbf{70.8} & \underline{57.1} & \underline{50.6} & \textbf{37.4} & \textbf{78.7} & \textbf{58.9} \\
\bottomrule
\end{tabular}
\end{table*}




\subsection{Main Results}
\label{sec:main_results}

Table~\ref{tab:main} summarizes the main results. We highlight the following key findings:

\paragraph{MIND-Skill leads consistently across diverse settings.}
When Qwen3.5-122B-A10B is used to generate skills, MIND-Skill achieves the highest TGC on both AppWorld splits and the highest BFCL-v3 accuracy, yielding the best average score (\textbf{59.1}), surpassing ACE (56.1) and Trace2Skill (55.1) by clear margins. Notably, on AppWorld-Challenge SGC, MIND-Skill significantly outperforms SOTA baselines (\textbf{39.6} vs.\ 34.5 for ACE and 33.1 for Trace2Skill).
We also note that no baseline performs consistently across both AppWorld splits: Trace2Skill scores higher than ACE on AppWorld-Normal TGC (67.3 vs.\ 65.5) but lower on AppWorld-Challenge TGC (46.8 vs.\ 51.1), suggesting that their generated skills may overfit to simpler task patterns.
MIND-Skill is the only method that leads on both splits simultaneously, and its large SGC advantage on AppWorld-Challenge indicates that the generated skills capture scenario-level procedural patterns rather than task-specific shortcuts.

\paragraph{Closed-loop optimization outperforms one-shot induction.}
Skill-extract uses the same induction agent as MIND-Skill to induce skills from trajectories in a single pass, isolating the contribution of our closed-loop optimization procedure (cf.~\S\ref{sec:deduction}--\S\ref{sec:optim}).
The gap is substantial: MIND-Skill outperforms Skill-extract by \textbf{8.1} and \textbf{7.2} on average when using Qwen3.5-122B-A10B and GPT-5.4 as the base model for the induction agent, respectively.
This confirms that one-shot skill extraction, even when the underlying induction agent is capable, cannot ensure the generated skills are faithful, generalizable, and well-structured without iterative optimization driven by our three textual losses.

\paragraph{Weak models match frontier ones with MIND-Skill.}
When skills are generated by GPT-5.4, MIND-Skill again achieves the highest average (\textbf{58.9}), outperforming Trace2Skill (56.6) and ACE (56.3).
On AppWorld-Challenge, Trace2Skill leads in terms of TGC, while MIND-Skill achieves the highest SGC (\textbf{37.4}) and leads on Normal TGC (\textbf{70.8}) as well as BFCL-v3 (\textbf{78.7}), showing our superiority across benchmarks.
An interesting observation is that MIND-Skill with the weaker Qwen3.5-122B-A10B as the base model for the induction agent achieves performance (\textbf{59.1} on average) comparable to MIND-Skill with GPT-5.4 (\textbf{58.9} on average).
This suggests that our induction-deduction framework can largely compensate for the capability gap between different base models, making high-quality skill generation accessible without relying on frontier models. We present a case study comparing the skills generated by Qwen3.5-122B-A10B and GPT-5.4 in Appendix~\ref{app:case_study}.

\subsection{Ablation Study and Further Analysis}
\label{sec:ablation}

\begin{figure*}[t]
  \centering
  \includegraphics[width=\textwidth]{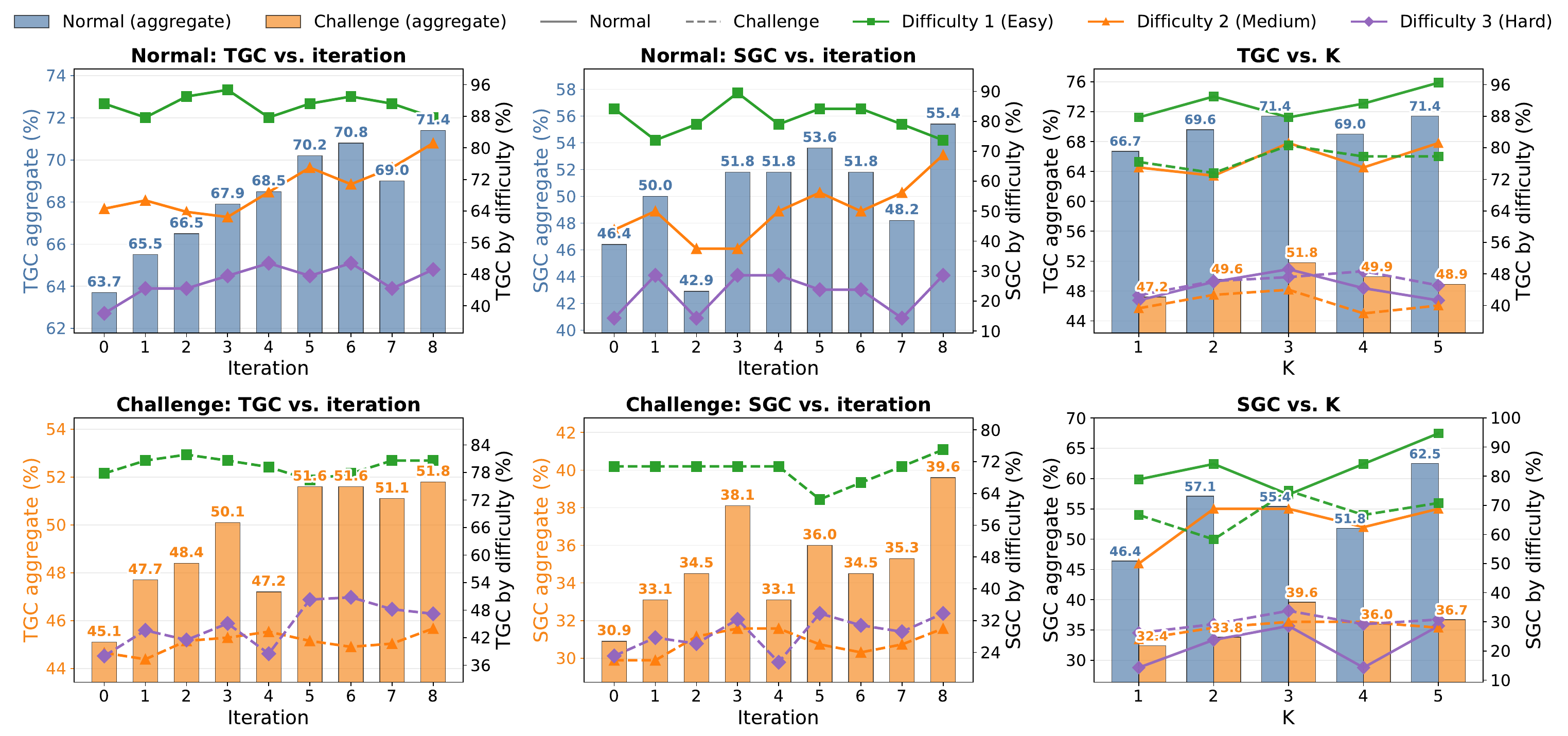}
  \vspace{-6mm}
  \caption{Performance at each iteration and the effect of varying the number of retrieved skills on AppWorld. Columns~1--2: TGC and SGC over iterations on Normal (top) and Challenge (bottom). Column~3: TGC and SGC across $K$, with both splits per panel. Bars (left axis) report the aggregate; lines (right axis) report per-difficulty accuracy.}
  \label{fig:combined-iter-k}
   \vspace{-6mm}

\end{figure*}

\begin{wrapfigure}{r}{0.45\textwidth}
\centering
\vspace{-4mm}
\includegraphics[width=0.45\textwidth]{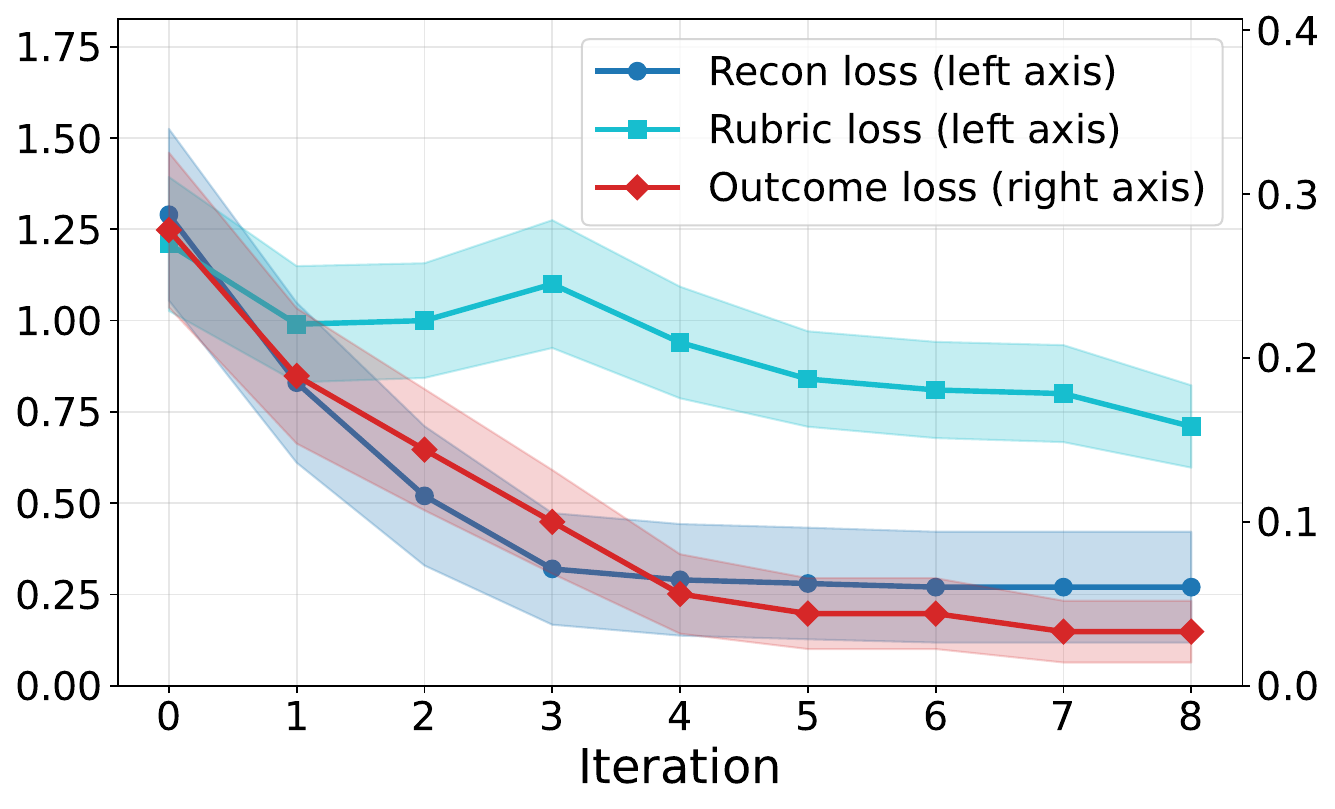}
\vspace{-6mm}
\caption{Loss values at each iteration on AppWorld. Shaded areas show ±1 SEM.}
\label{fig:loss}
\vspace{-4mm}
\end{wrapfigure}

\paragraph{Skill quality improves steadily across optimization iterations.}
Figure~\ref{fig:combined-iter-k} (columns~1--2) tracks test performance across optimization iterations. At each iteration, each task's skill library entry is updated to the best skill found so far via the lexicographic selection described in Algorithm~\ref{alg:MIND-Skill} (lines~12--13). Starting from iteration~0, which is equivalent to Skill-extract, TGC improves by \textbf{7.7} on Normal and by \textbf{6.7} on Challenge over 8 iterations, with the majority of gains concentrated in the first 3 rounds. Per-difficulty breakdowns show that easy tasks saturate early while hard tasks continue to benefit from later iterations, suggesting that early rounds fix coarse procedural gaps whereas later rounds resolve subtler edge cases. Figure~\ref{fig:loss} confirms this dynamic: all three losses decrease steadily with small variance, and the outcome loss drops to near zero within 3 iterations while the reconstruction and rubric losses continue to decrease in later iterations.

\paragraph{The effect of varying the number of retrieved skills $K$.}
Figure~\ref{fig:combined-iter-k} (column~3) shows the effect of varying $K$ injected at inference time. $K{=}1$ underperforms across all metrics, as a single skill may not cover the full procedural scope of a test task. Performance improves substantially from $K{=}1$ to $K{=}3$, as retrieving multiple complementary skills broadens procedural coverage and reduces the agent's sensitivity to any single poor match. Per-difficulty breakdowns confirm this: easy tasks are near-ceiling from $K{\ge}2$, while medium and hard tasks benefit most from the additional coverage at $K{=}3$. $K{=}5$ pushes Normal SGC further to 62.5, indicating that more skills can still help with scenario-level consistency. Balancing overall performance, we use $K{=}3$ for all main experiments.

\begin{wraptable}{r}{0.52\textwidth}
\vspace{-4mm}
\centering
\small
\setlength{\tabcolsep}{6pt}
\caption{Ablation study on AppWorld.}
\label{tab:ablation}
\vspace{2mm}
\begin{tabular}{l cc cc}
\toprule
& \multicolumn{2}{c}{\textbf{Normal}} & \multicolumn{2}{c}{\textbf{Challenge}} \\
\cmidrule(lr){2-3} \cmidrule(lr){4-5}
\textbf{Method} & TGC & SGC & TGC & SGC \\
\midrule
Skill-extract       & 63.7 & 46.4 & 45.1 & 30.9 \\
\midrule
w/o outcome         & 68.5 & 55.4 & 48.0 & 34.5 \\
w/o reconstruction  & 66.1 & 46.4 & 45.8 & 31.6 \\
w/o rubric          & 64.3 & 46.4 & 46.5 & 31.6 \\
MIND-Skill (full)   & \textbf{71.4} & \textbf{55.4} & \textbf{51.8} & \textbf{39.6} \\
\bottomrule
\end{tabular}
\end{wraptable}

\paragraph{Each loss component contributes to skill quality.}
Table~\ref{tab:ablation} ablates each loss function on AppWorld.
All ablated variants outperform Skill-extract, confirming that each loss is indispensable for high-quality skill generation.
Removing the reconstruction loss causes the largest Challenge TGC drop (\textbf{51.8} $\to$ 45.8), nearly erasing all gains over Skill-extract (45.1). Without comparing reconstructed and source trajectories, the optimizer lacks the fine-grained procedural feedback needed to identify missing key steps and flawed workflows. Removing the rubric loss causes the largest Normal TGC drop (\textbf{71.4} $\to$ 64.3). Without abstraction-level regularization, the optimizer tends to leak instance-specific details into skills, which may coincidentally help on certain challenge tasks but hurt generalization across the broader task population.
Removing the outcome loss has the mildest effect. Notably, even without any ground-truth execution feedback, the w/o outcome variant (\textbf{68.5} Normal TGC, \textbf{48.0} Challenge TGC) already outperforms Trace2Skill on both splits. This highlights that the reconstruction and rubric losses alone provide sufficiently rich signal to surpass concurrent trajectory-distillation methods. Nonetheless, outcome loss catches runtime errors and silent API failures that textual judgment alone misses, helping the full MIND-Skill improve Challenge TGC to \textbf{51.8} and SGC to \textbf{39.6}.

\begin{wrapfigure}{r}{0.45\textwidth}
\centering
\vspace{-3mm}
\includegraphics[width=0.45\textwidth]{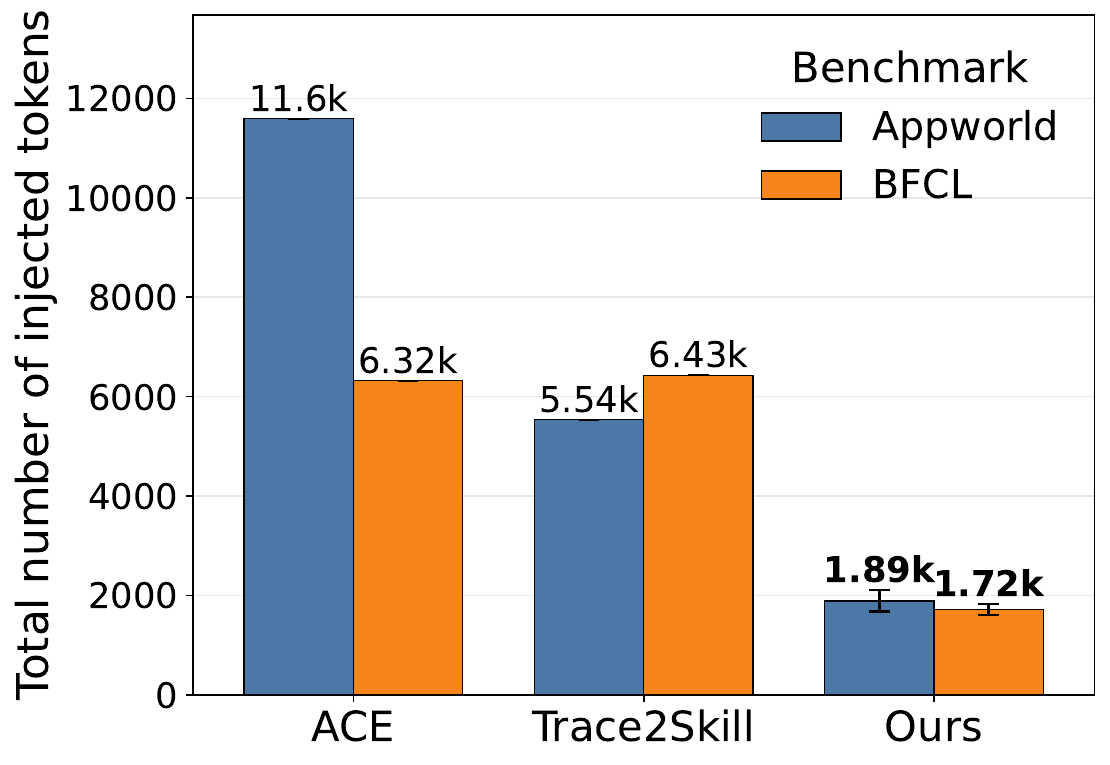}
\vspace{-5mm}
\caption{Total number of injected tokens.}
\label{fig:skill_size}
\vspace{-2mm}
\end{wrapfigure}

\paragraph{MIND-Skill generates compact skills.}
Figure~\ref{fig:skill_size} compares the total number of injected tokens at inference time. Although MIND-Skill retrieves $K{=}3$ skills per test task, the number of injected tokens remains $3{-}6{\times}$ smaller than ACE's monolithic playbook and Trace2Skill's single skill directory. Our rubric loss explicitly penalizes redundant boilerplate, encouraging the optimizer to retain only essential procedural content. In contrast, ACE and Trace2Skill pack all training-time knowledge into a single monolithic artifact regardless of task relevance. MIND-Skill instead follows a progressive-disclosure principle, where each retrieved skill covers only the procedural knowledge relevant to its matched task category. This produces compact yet actionable skills without sacrificing effectiveness.

\section{Conclusion}
\label{sec:conclusion}
In this work, we presented \textbf{MIND-Skill}, a multi-agent induction and deduction framework for automatically synthesizing high-quality agent skills from successful execution trajectories. MIND-Skill departs from prior skill-generation approaches by introducing a closed-loop process that explicitly validates and refines generated skills through trajectory reconstruction, execution feedback, and comprehensive rubric assessment. Specifically, MIND-Skill combines an induction agent and a frozen deduction agent with three complementary textual losses: reconstruction loss, outcome loss, and rubric loss. These losses are jointly optimized with TextGrad to iteratively refine the induction prompt, improving generated skills in terms of faithfulness, task correctness, and documentation quality. Experiments on AppWorld and BFCL-v3 show that the resulting skills improve agent performance on both source tasks and held-out tasks unseen during skill generation, demonstrating the effectiveness and generalizability of the proposed framework.


\bibliographystyle{plainnat}
\bibliography{references}

\newpage
\appendix

\section{Limitations and Broader Impacts}
\label{sec:limitations}

MIND-Skill requires successful trajectories as input to the skill
induction process, since the reconstruction loss relies on a reference trajectory to provide optimization signals. This couples the scope of the generated skill library to the set of tasks for which successful trajectories can be obtained. In practice, however,
this dependency can be satisfied in multiple ways: besides model rollouts,
ground-truth solution scripts can also serve as surrogate trajectories,
as described in our fallback strategy (Appendix~\ref{app:mind_skill_impl}). 

MIND-Skill aims to automate the creation of reusable agent skills, reducing the manual effort required from domain experts and making high-quality procedural knowledge more accessible. Our results show that even weaker models can produce competitive skills after optimization, which could democratize access to capable AI agents. On the other hand, as with any advance in autonomous agent capabilities, automatically generated skills could in principle be used to automate undesirable agent behaviors. We note that this risk is shared broadly across the agent skill and agent framework literature and is not specific to our method. The skills produced by MIND-Skill are human-readable Markdown documents, which facilitates auditing and oversight before deployment.

\section{Implementation Details}
\label{app:details}

All experiments use Qwen3.5-122B-A10B with extended thinking disabled as the inference model. In the cross-model setting (Table~\ref{tab:main}, lower block), GPT-5.4 is used for skill generation and optimization while inference remains on Qwen; the exact role assignments per method are detailed below. All LLM calls are issued through OpenRouter.\footnote{\url{https://openrouter.ai/}} For every Qwen-3.5 call we additionally pin the upstream provider to the model's native vendor Alibaba via OpenRouter's \texttt{provider} routing field,\footnote{\url{https://openrouter.ai/docs/features/provider-routing}} because open-weight models on OpenRouter are served by multiple upstream providers (e.g., Alibaba, Novita, AtlasCloud, Venice) whose deterministic batching and CUDA kernels differ enough to produce cross-provider variance. All methods share the same training--test partition.

\subsection{ACE Implementation Details}
\label{app:ace}
We re-implement ACE on AppWorld and BFCL following the official released code, using Qwen-122B for all three roles (Generator, Reflector, Curator). In the cross-model setting, the Reflector and Curator are replaced with GPT-5.4; the Generator and inference agent remain on Qwen-122B. We use the offline-with-GT mode, which is the most directly analogous setting to our training pipeline: both use ground-truth solutions during training. Training proceeds sequentially over the training split: for each task, the Generator produces a ReAct trajectory; if the trajectory fails its unit test, the Reflector compares it against the ground-truth solution code to diagnose the failure; the Curator then distills the reflection into structured bullets (e.g., strategies and hard rules, API usage patterns, common mistakes) that are appended to a shared playbook. We allow up to 5 retries per training task following the released configuration. At test time, the entire accumulated playbook is injected verbatim into the agent's system prompt regardless of task relevance.

\subsection{Trace2Skill Implementation Details}
\label{app:trace2skill}
We re-implement Trace2Skill on AppWorld and BFCL following the open-source release (Skill Creation mode), using Qwen-122B for all roles. In the cross-model setting, the success/error analysts and the hierarchical merge LLM are replaced with GPT-5.4; rollout collection and inference remain on Qwen-122B. This is by Trace2Skill's design: the analysts must observe successes and failures of the same model that will be deployed at test time, so the resulting skill targets its failure modes. We roll out on the training tasks. The success analyst processes each passing trajectory in a single LLM call. The error analyst runs an agentic ReAct loop (max\_turns=40) with pass-gating; since AppWorld's evaluation depends on cumulative database state rather than the single-shot script-vs-file comparison assumed by the original code, we replace the evaluator with a stateful REPL exposing \texttt{appworld\_execute}, \texttt{appworld\_evaluate}, and \texttt{appworld\_reset}. All other methodology (causality gate, hierarchical merge, patch vocabulary) is preserved. We run the hierarchical merge pipeline, and the resulting skill directory is injected verbatim into the ReAct system prompt at inference time.

\subsection{MIND-Skill Implementation Details}
\label{app:mind_skill_impl}
Qwen-122B is used as the base model for every agent (induction, deduction, judges, gradient, and optimizer) in the self-generation setting. In the cross-model setting, GPT-5.4 replaces all roles except deduction and inference, which remain on Qwen-122B. The ablation studies and analyses in \S\ref{sec:ablation} use the self-generation setting with Qwen-122B as the induction agent. For each training task, we obtain a reference trajectory by rolling out strong model in the AppWorld environment; if the rollout fails the task checker, we fall back to wrapping the ground-truth solution code as a trajectory. After optimization ($Q{=}8$ iterations per task), we retain the best-so-far skill for each task (Algorithm~\ref{alg:MIND-Skill}, lines~12--13), yielding a library with one skill per training task. At test time, only each skill's \texttt{name} and \texttt{description} are exposed to a Qwen-122B retrieval call that selects the top-$K$ most relevant skills; their full markdown bodies are concatenated into the agent's skill slot (\S\ref{app:decoder_prompt}). We set $K{=}3$ as the default and report ablations over $K \in \{1,2,3,4,5\}$. Every LLM call is wrapped in a 3-attempt retry: empty responses (rate limits, transient outages) trigger a same-message retry, while non-empty responses that fail schema validation are appended back as an assistant turn followed by a fix instruction, recovering both API failures and format violations. 

\newpage

\section{Case Studies}
\label{app:case_studies}

We present two case studies that complement the quantitative results. The first (\S\ref{app:skill_quality}) examines why baseline-generated skills suffer from quality issues that MIND-Skill avoids. The second (\S\ref{app:case_study}) investigates why skills generated by a weaker model can match those from a stronger one.

\begin{figure*}[t]
\centering
\begin{minipage}[t]{0.48\textwidth}
{\small\textbf{(a) ACE: task-specific memorization}}
\vspace{1mm}
\begin{tcolorbox}[colback=contextbg, colframe=boxborder,
  boxsep=1pt, left=3pt, right=3pt, top=2pt, bottom=2pt,
  arc=1pt, boxrule=0.5pt]
{\scriptsize
\textbf{Task 60d0b5b:} \texttt{``The last Venmo payment request I sent to Robert was an accident and they approved it. Send them the money back.''}
}
\end{tcolorbox}
\vspace{0.5mm}
\begin{tcolorbox}[colback=boxbg, colframe=boxborder,
  boxsep=1pt, left=3pt, right=3pt, top=2pt, bottom=2pt,
  arc=1pt, boxrule=0.5pt]
{\scriptsize\ttfamily
\textcolor{gray!60}{\# Section: STRATEGIES AND HARD RULES}\\[1pt]
Venmo Refund Workflow: When a task requires\\
refunding an accidental `payment request'\\
that was approved:\\
\colorbox{hlblue}{(1) use show\_sent\_payment\_requests}\\
\colorbox{hlblue}{\quad\; with status=`approved'}\\
\colorbox{hlblue}{(2) filter by receiver's email}\\
\colorbox{hlblue}{(3) sort by created\_at descending}\\
\colorbox{hlblue}{(4) take the most recent request}\\
\colorbox{hlblue}{(5) refund via create\_transaction}
}
\end{tcolorbox}
\end{minipage}
\hfill
\begin{minipage}[t]{0.48\textwidth}
{\small\textbf{(b) Trace2Skill: malformed structure}}
\vspace{1mm}
\begin{tcolorbox}[colback=boxbg, colframe=boxborder,
  boxsep=1pt, left=3pt, right=3pt, top=2pt, bottom=2pt,
  arc=1pt, boxrule=0.5pt]
{\scriptsize\ttfamily
\#\# When to Apply\\
- Filesystem-style tasks: navigating ...\\
- Social / messaging tasks: posting ...\\
- Booking / ticketing / trading: creating ...\\
- Multi-domain composition: tasks ...\\
\textcolor{gray!60}{(2 more applicability bullets)}\\[1pt]
\colorbox{hlblue}{3a. Trust tool responses: Treat all}\\
\colorbox{hlblue}{\quad\; values returned by tools ...}\\
\colorbox{hlblue}{3b. Calculate derived values and ...}\\
\textcolor{gray!60}{(3c--3m: 10 more procedural steps)}\\
\colorbox{hlblue}{3m. ... pass them as a list argument}\\
\colorbox{hlorange}{\quad\; (e.g., [5000, 7000]).\#\# Procedure}\\[1pt]
1.\;\; Per-turn intent extraction: Read ...\\
1.5\; Check initial service state: Before ...\\
2.\;\; Schema check before emitting: For...
}
\end{tcolorbox}
\end{minipage}

\vspace{2mm}
\begin{minipage}[t]{0.98\textwidth}
{\small\textbf{(c) MIND-Skill: transferable and well-structured }}
\vspace{1mm}
\begin{tcolorbox}[colback=boxbg, colframe=boxborder,
  boxsep=1pt, left=3pt, right=3pt, top=2pt, bottom=2pt,
  arc=1pt, boxrule=0.5pt]
{\scriptsize\ttfamily
\begin{minipage}[t]{0.30\textwidth}
\#\# When to Apply\\
- Target API requires auth token via a separate login step.\\

- List endpoint returns data in\\
\quad\; pages, requiring a loop.\\

- Task involves identifying one\\
\quad\; item by a \colorbox{hlgreen}{label or attribute},\\
\quad\; modifying it, and applying a\\
\quad\; different state change to all\\
\quad\; \colorbox{hlgreen}{remaining items}.
\end{minipage}
\hfill
\begin{minipage}[t]{0.30\textwidth}
\#\# Procedure\\
1. Authenticate: Retrieve token.\\
2. Paginate and Collect: Loop until empty to collect all.\\
3. Identify Target: Find item by \colorbox{hlgreen}{specific label}.\\
4. Apply Specific Update:\\
\quad\; Modify target (e.g., \colorbox{hlgreen}{time shift}).\\
5. Apply Bulk Update:\\
\quad\; \colorbox{hlgreen}{Disable} the rest.\\
6. Verify: Re-fetch and confirm.
\end{minipage}
\hfill
\begin{minipage}[t]{0.30\textwidth}
\#\# Key Patterns\\
- Pagination Loop: while-loop,\\
\quad\; break on empty page.\\
- Selective Mutation: unique\\
\quad\; update for target, uniform\\
\quad\; for rest.\\
- State Verification: post-update re-fetch to validate.\\[3pt]
\#\# Common Pitfalls\\
- Failing to loop through all\\
\quad\; pages, missing items.\\
- Updating target incorrectly\\
\quad\; or failing to exclude it\\
\quad\; from the bulk update.
\end{minipage}
}
\end{tcolorbox}
\end{minipage}

\caption{\textbf{(a)}~ACE encodes the solution of training task \texttt{60d0b5b} as a five-step recipe (\colorbox{hlblue}{\scriptsize blue}: memorized steps); its trigger is a near-verbatim paraphrase of the task instruction.
\textbf{(b)}~Trace2Skill misplaces 13 procedural steps (\colorbox{hlblue}{\scriptsize blue}) under ``When to Apply'' and concatenates a section header inline (\colorbox{hlorange}{\scriptsize orange}).
\textbf{(c)}~MIND-Skill uses only conceptual placeholders (\colorbox{hlgreen}{\scriptsize green}: e.g., ``label or attribute'', ``time shift'') instead of memorized app-specific values, and maintains clean section boundaries throughout.}
\label{fig:rubric_examples}
\end{figure*}

\subsection{Case Study: Skill Quality Degrades Without Explicit Guarantees}
\label{app:skill_quality}
Figure~\ref{fig:rubric_examples} contrasts representative skills from all three methods. ACE's playbook entry~(a) memorizes the solution of a single training task as a high-priority rule, which misfires on any test task that deviates from that scenario. Trace2Skill's SKILL.md~(b) misplaces procedural steps under the wrong section header, producing malformed structure that downstream agents struggle to parse. In contrast, the MIND-Skill entry~(c), generated from training task \texttt{302c169\_1}, uses only conceptual placeholders (``label or attribute'', ``time shift'', ``remaining items'') rather than memorized app-specific values, and maintains clean section boundaries where Procedure contains only ordered actions, Key Patterns names only transferable abstractions, and Common Pitfalls lists only failure modes. These differences directly reflect our rubric loss: its \textit{ground-truth independence} dimension penalizes task-specific memorization as in~(a), its \textit{actionability} and \textit{completeness} dimensions enforce structural coherence absent in~(b), and together they produce skills like~(c) that are both transferable and well-structured.

\begin{figure*}[t]
\centering
\begin{minipage}[t]{0.48\textwidth}
{\small\textbf{(a) Qwen-self skill (Net +12)}}
\vspace{1mm}
\begin{tcolorbox}[colback=boxbg, colframe=boxborder,
  boxsep=1pt, left=3pt, right=3pt, top=2pt, bottom=2pt,
  arc=1pt, boxrule=0.5pt]
{\scriptsize\ttfamily
\#\# Procedure\\
1. Authenticate: Obtain access token.\\
2. Paginate and Collect: Loop pages\\
\quad\; until empty to collect all items.\\
3. Identify Target: Find item by label.\\
4. Apply Specific Update: Modify target.\\
5. Apply Bulk Update: Disable the rest.\\
6. Verify: Re-fetch and confirm.\\[3pt]
\#\# Key Patterns\\
\colorbox{hlblue}{- Pagination Loop: while-loop, break}\\
\colorbox{hlblue}{\quad\; on empty page.}\\
\colorbox{hlblue}{- Selective Mutation: unique update}\\
\colorbox{hlblue}{\quad\; for target, uniform for rest.}\\
\colorbox{hlblue}{- State Verification: post-update}\\
\colorbox{hlblue}{\quad\; re-fetch to validate.}
}
\end{tcolorbox}
\end{minipage}
\hfill
\begin{minipage}[t]{0.48\textwidth}
{\small\textbf{(b) GPT-teach skill (Net +1)}}
\vspace{1mm}
\begin{tcolorbox}[colback=boxbg, colframe=boxborder,
  boxsep=1pt, left=3pt, right=3pt, top=2pt, bottom=2pt,
  arc=1pt, boxrule=0.5pt]
{\scriptsize\ttfamily
\#\# Procedure\\
1. Inspect API docs for endpoints.\\
2. Authenticate and store token.\\
3. Read listing endpoint docs.\\
4. Paginate until empty, collect all.\\
5. Identify target by attribute.\\
6. Read update endpoint docs.\\
7. Update target with modification.\\
8. Bulk-update all non-target items.\\
9. Re-fetch and verify both conditions.\\
10. Mark task complete.\\[3pt]
\#\# Key Patterns\\
\colorbox{hlorange}{- Doc-first execution}\\
\colorbox{hlorange}{- Credential bootstrap}\\
\colorbox{hlorange}{- Paginate-until-empty}\\
\colorbox{hlorange}{- Target-then-bulk}\\
\colorbox{hlorange}{- Verify-by-refetch}
}
\end{tcolorbox}
\end{minipage}
\caption{Paired skills from the same training task (\texttt{302c169\_1}). Net contribution = test tasks flipped from fail to pass minus pass to fail, relative to the no-skill baseline. Both skills encode the same procedural logic, but differ in vocabulary: Qwen-self uses plain labels (\colorbox{hlblue}{\scriptsize blue}) while GPT-teach adopts textbook-style pattern names (\colorbox{hlorange}{\scriptsize orange}).}
\label{fig:case_study}
\end{figure*}

\subsection{Case Study: Why Weaker Skill Generators Can Match Stronger Ones}
\label{app:case_study}

A perhaps counterintuitive finding in Table~\ref{tab:main} is that MIND-Skill with the weaker Qwen3.5-122B-A10B as skill generator (59.1 average) achieves comparable performance to MIND-Skill with GPT-5.4 (58.9 average). We investigate this through a paired case study on training task \texttt{302c169\_1}, where both pipelines optimize a skill from the same source trajectory. We measure each skill's \textit{net contribution} by tracking all test tasks that retrieved it: among those tasks, we count how many flipped from fail to pass after skill injection, minus how many regressed from pass to fail, relative to the no-skill baseline. The Qwen-self skill achieves a net contribution of +12, while the GPT-teach skill achieves only +1.

Figure~\ref{fig:case_study} compares the two skills side by side. Both encode the same procedural logic (authenticate, paginate, identify, update, verify), yet they differ markedly in style. The GPT skill adopts textbook-style pattern names (``Doc-first execution'', ``Credential bootstrap'', ``Target-then-bulk'') and includes defensive caveats (``check partial-update behavior'', ``confirm paging behavior and returned attributes'') that reflect GPT-5.4's own reasoning preferences. The Qwen skill uses plainer vocabulary (``Pagination Loop'', ``Selective Mutation'') at the abstraction level Qwen naturally operates at. When injected into Qwen's prompt at inference time, the self-authored skill is decoded naturally, whereas the GPT-authored skill requires implicit style adaptation that can dilute the procedural signal. This is not a matter of correctness; GPT's labels are arguably more precise, but precision in a foreign dialect does not help the inference model act on it. Moreover, the GPT skill is calibrated to its own capability: it recommends fine-grained checks (e.g., inspecting partial-update semantics) that GPT-5.4 can execute but Qwen cannot operationalize, consuming attention on sophistication the inference model has no headroom to exploit. Self-training avoids both costs by construction, since the writer and reader share the same distribution and capability profile. This suggests that after quality-guaranteed optimization, such alignment becomes a more important factor than the raw reasoning capability of the skill generator.

\newpage

\section{Prompt Design}
\label{app:prompts}

This section presents the key prompts used in MIND-Skill. For readability, all prompts are abbreviated to their essential structure and instructions.

\subsection{Induction Prompt and Its Evolution}
\label{app:induction_prompt}

The induction agent's system prompt $\mathcal{P}_I$ is the sole variable optimized by TextGrad. Figure~\ref{fig:prompt_evolution} contrasts the universal initial prompt $\mathcal{P}_I^{(0)}$ with an optimized variant $\mathcal{P}_I^{*}$ obtained after four iterations on a representative training task. The optimizer inserts domain-pattern rules and explicit abstraction-leakage warnings with paired Bad/Good examples, growing the prompt from ${\sim}$530 to ${\sim}$2.0K tokens. These additions are textual rules derived from gradient feedback on specific failure modes, not manual engineering.

\begin{figure*}[t]
\centering
\begin{minipage}[t]{0.48\textwidth}
\begin{promptbox}{Initial Induction Prompt $\mathcal{P}_I^{(0)}$ \normalfont{\scriptsize($\sim$530 tokens)}}
{\scriptsize
\textbf{Role:} You are an expert at extracting reusable procedural strategies from task solutions.

Given a task instruction and its solution code, extract a SKILL that describes the procedure pattern---the structural ``how-to'' that is NOT obvious from the instruction alone.

\textbf{Rules for a good skill:}
\begin{enumerate}[leftmargin=*, nosep]
\item Describe ONLY solving strategy and structural patterns: authentication flow, pagination/iteration, multi-step data retrieval, data transformation, output construction.
\item Do NOT include task-specific info: no specific API names, field names, entity names, thresholds. \textit{Test: if someone can guess the original task from your skill alone, it is too specific.}
\item Focus on NON-OBVIOUS structural knowledge.
\end{enumerate}

\textbf{Output:} Valid SKILL.md with YAML frontmatter, followed by sections: Overview, When to Apply, Procedure, Key Patterns, Common Pitfalls.
}
\end{promptbox}
\end{minipage}
\hfill
\begin{minipage}[t]{0.48\textwidth}
\begin{promptbox}{Optimized Prompt $\mathcal{P}_I^{*}$ \normalfont{\scriptsize($\sim$2.0K tokens, iteration~5)}}
{\scriptsize
\textit{[Role and Rule 1 prefix unchanged]}

\colorbox{hlblue}{\textbf{+ Identifier \& Scope Resolution:}} Discover the correct data source via parent-container endpoint; inspect schema for the unique key field.

\colorbox{hlblue}{\textbf{+ Constraint-Based Data Filtering:}} Construct subsets by hierarchy traversal; warn that a global ``list all'' may exceed the task scope.

\colorbox{hlblue}{\textbf{+ Output Construction \& Completion Signals:}} Silent completion when no report is requested; describe structural schema (not values) when a report IS requested.

\textit{[Rule 2 expanded with:]}

\colorbox{hlblue}{\textbf{+ STRICT ABSTRACTION RULE:}} Never include specific endpoints, field names, or boolean values.\\
\textbf{Bad:} ``Call \texttt{show\_song\_privates} to check if \texttt{liked} is true.''\\
\textbf{Good:} ``Call the endpoint that exposes user-specific state flags.''

\textit{[+ Common Pitfalls scaffold with solution rules for: hallucinating endpoints, assuming global lists match constraints, parameter consistency.]}
}
\end{promptbox}
\end{minipage}
\caption{The induction agent's system prompt is the sole variable optimized by TextGrad. \textbf{(a)}~The universal initial prompt $\mathcal{P}_I^{(0)}$ used for all training tasks. \textbf{(b)}~The optimized prompt $\mathcal{P}_I^{*}$ after 4 TextGrad iterations on source task \texttt{692c77d\_1}. Highlighted spans (\colorbox{hlblue}{\scriptsize blue}) are rules TextGrad inserted to address failure modes observed during training, including the explicit Bad/Good examples for the abstraction rule.}
\label{fig:prompt_evolution}
\end{figure*}

\newpage

\subsection{Deduction Agent}
\label{app:decoder_prompt}

Both training and evaluation share the same ReAct template, differing only in the content injected into the skill slot. Figure~\ref{fig:decoder_prompt} illustrates the template structure and the injection mechanism. The template consists of: (i) framing prose orienting the agent to AppWorld's API discovery tools, (ii) the skill injection slot between \texttt{SKILLS BEGIN/END} markers, (iii) in-context ReAct demonstration trajectories, and (iv) the real task instruction. All baselines in our comparison (ACE, Trace2Skill) share this same template and differ only in what fills the skill slot. At training time, the slot receives one candidate skill being optimized; at evaluation time, it receives $K$ retrieved skills.

\begin{figure}[t]
\begin{promptbox}{Deduction Agent: Template Structure}
{\small
\texttt{USER:}\\
\texttt{I am your supervisor and you are a super intelligent AI assistant whose job is to achieve my tasks autonomously.}\\
\texttt{To do this, you will interact with apps using their APIs ...}\\
{\scriptsize\textcolor{gray}{[API discovery orientation: \texttt{show\_app\_descriptions()}, \texttt{show\_api\_doc()}, etc.]}}

\texttt{You are also provided with a curated set of skills to help you solve the task effectively.}\\
\texttt{Read the skills first, then execute the task by explicitly leveraging each relevant section:}

\texttt{\#\#\# SKILLS BEGIN}\\
\texttt{\{\{ skills \}\}} \hfill{\scriptsize\textcolor{gray}{training: 1 skill $\mid$ evaluation: $K$ retrieved skills}}\\
\texttt{\#\#\# SKILLS END}

{\scriptsize\textcolor{gray}{[in-context ReAct demonstrations]}}

\texttt{USER:}\\
\texttt{My name is: \{\{first\_name\}\} \{\{last\_name\}\}.}\\
\texttt{Task: \{\{ input\_str \}\}} \hfill{\scriptsize\textcolor{gray}{real task instruction}}
}
\end{promptbox}
\caption{Abbreviated structure of the deduction agent's prompt template. Skills enter through a single \texttt{\{\{skills\}\}} slot; the only difference between training and evaluation is the number of injected skills (1 vs.\ $K$).}
\label{fig:decoder_prompt}
\end{figure}

\newpage

\subsection{Textual Loss Prompts}
\label{app:loss_prompts}

Our three textual losses are implemented as LLM judge calls that return scores on a 0--10 scale (higher is better). We adopt this convention because LLMs produce more calibrated assessments when prompted to score quality directly---for instance, a rubric score of 8/10 carries clear semantic meaning, whereas a loss value of 2 lacks intuitive grounding. To conform to the standard minimization convention, we convert scores to losses via $\ell = c - \text{score}$, where $c$ is the upper bound of the scoring range. Figure~\ref{fig:rubric_prompt} presents the rubric loss prompt, which instructs the judge to classify each claim in the skill along a GT-leakage counterfactual and score five quality dimensions. Figure~\ref{fig:traj_judge_prompt} presents the reconstruction loss prompt, which evaluates procedural alignment between the source and reconstructed trajectories. The outcome loss requires no prompt as it is computed directly from environment execution results.

\begin{figure}[t]
\begin{promptbox}{Skill Quality Rubric Prompt (\texttt{RUBRIC\_SYSTEM})}
{\small
\textbf{Role:} You judge the quality of a procedural skill that will later be used by a coder who has NOT seen the reference solution.

\textbf{You are given:}
\begin{enumerate}[leftmargin=*, nosep, label=(\arabic*)]
  \item The task instruction (the same one the coder will see)
  \item The skill that was extracted from the (hidden) reference solution
\end{enumerate}

\textbf{The Central Question:} How much of this skill is useful procedural knowledge vs.\ leaked solution details?

\textbf{For each claim in the skill, classify it:}
\begin{enumerate}[leftmargin=*, nosep, label=(\Alph*)]
  \item \textbf{Standard convention} --- a general software pattern the coder can assume without context.
  \item \textbf{Inferable from instruction} --- derivable from the task text alone.
  \item \textbf{Leaked from ground truth} --- unknowable without seeing the reference solution (exact API paths, specific algorithm choices, library decisions, hard-coded thresholds).
\end{enumerate}

\textbf{Score each dimension (0--10):}
\begin{enumerate}[leftmargin=*, nosep]
  \item \textbf{GT-Independence}: fraction of content a developer could write from instruction alone.
  \item \textbf{Actionability}: can a coder use this + API docs to write working code?
  \item \textbf{Transferability}: would this apply to a structurally similar task in a different domain?
  \item \textbf{Completeness}: full procedure chain covered?
  \item \textbf{Conciseness}: information-dense, no redundant boilerplate?
\end{enumerate}

\textbf{Output:} JSON with per-dimension scores, leaked claims, and issue summary.
}
\end{promptbox}
\caption{Five-axis rubric prompt with GT-leakage counterfactual. The overall score is gated on GT-independence to prevent overfit skills from masquerading as actionable.}
\label{fig:rubric_prompt}
\end{figure}

\begin{figure}[t]
\begin{promptbox}{Trajectory Reconstruction Judge Prompt (\texttt{TRAJECTORY\_JUDGE\_SYSTEM})}
{\small
\textbf{Role:} Evaluate whether a deduction agent's ReAct trajectory follows the same procedural strategy as a reference trajectory on the same task.

\textbf{Evaluate on procedural alignment, not literal text match.}

\textbf{Criteria:}
\begin{enumerate}[leftmargin=*, nosep]
  \item Same sequence of API-call families (e.g., auth $\to$ list $\to$ detail $\to$ action), order and dependencies.
  \item Same control flow: pagination, accumulation loops, early-exit conditions, branching.
  \item Deduction agent's final environment observation converges to the same outcome.
\end{enumerate}

\textbf{Ignore:} Different variable names, intermediate print statements, step-count differences, specific IDs/values.

\textbf{Output:} JSON with alignment score (0--10), boolean flags for API sequence / control flow / final state match, and list of procedural mismatches.
}
\end{promptbox}
\caption{Trajectory reconstruction judge prompt. Scores procedural alignment rather than literal text match; tolerates step-count and variable-name differences as long as API-family sequence and control flow agree.}
\label{fig:traj_judge_prompt}
\end{figure}

\newpage

\subsection{Gradient and Optimizer Prompts}
\label{app:grad_opt_prompts}

The gradient and optimizer LLMs form the two-step TextGrad update cycle. The gradient LLM (Figure~\ref{fig:gradient_prompt}) diagnoses failure patterns from rollout cases and produces textual feedback. The optimizer LLM (Figure~\ref{fig:optimizer_prompt}) then takes the current prompt and this feedback to produce an updated prompt, without access to rollout cases or scores. This separation ensures a clean diagnostic-then-apply workflow. Two design choices are worth noting: the gradient prompt explicitly instructs the LLM to refuse the naive fix of writing ground-truth-specific tokens into the skill even when execution failures seem to call for it, preserving GT-independence as a hard constraint; the optimizer prompt enforces a format-preservation rule to prevent the optimizer from deleting the SKILL.md output specification across iterations.

\begin{figure}[t]
\begin{promptbox}{Gradient LLM System Prompt}
{\small
\textbf{Role:} You are part of an optimization system that improves an induction agent prompt. The induction agent extracts procedural skills from task solutions. A deduction agent then uses these skills to reconstruct the solution.

\textbf{Your job:} Analyze cases where quality was low, and give feedback on how to improve the induction agent prompt so it produces better skills.

\textbf{Each case may include:}
\begin{enumerate}[leftmargin=*, nosep]
\item \textbf{Rubric scores} (0--10): GT-independence, actionability, transferability, conciseness, plus specific leaked claims flagged by the judge.
\item \textbf{Reconstruction score/issues}: deduction agent's trajectory vs.\ reference---shows what the deduction agent got wrong.
\item \textbf{Execution result}: pass/fail with error messages.
\end{enumerate}

\textbf{Critical tradeoff:} When execution fails because the deduction agent guessed the wrong field name, the naive fix is to write the correct name into the skill. \textbf{Do NOT endorse this.} It makes execution pass but destroys GT-independence. Better fix: improve the procedure wording so it guides the deduction agent to inspect API docs for the correct field, rather than hard-coding the field path.

\textbf{Output:} Describe what to change in the induction agent prompt and why. Do NOT propose a new prompt.
}
\end{promptbox}
\caption{Gradient LLM prompt. The LLM receives low-quality and high-quality rollout cases and produces a textual diagnosis. It is explicitly instructed to reject the naive fix of leaking ground-truth details into skills.}
\label{fig:gradient_prompt}
\end{figure}

\begin{figure}[t]
\begin{promptbox}{Optimizer LLM System Prompt}
{\small
\textbf{Role:} You are part of an optimization system that improves text prompts. You will receive the current prompt and feedback on its weaknesses. Produce an improved version.

\textbf{Rules:}
\begin{enumerate}[leftmargin=*, nosep]
\item Make targeted changes that address the specific feedback.
\item Do not break things that are already working.
\item Keep roughly the same length and structure.
\item \textbf{Hard constraint:} The prompt MUST keep the SKILL.md output format specification intact, including YAML frontmatter with \texttt{name} and \texttt{description} fields. Do not remove, weaken, or omit the format specification.
\end{enumerate}

\textbf{Output:} The improved prompt wrapped in \texttt{<IMPROVED\_VARIABLE>} tags. No explanation outside the tags.
}
\end{promptbox}
\caption{Optimizer LLM prompt receives the current prompt and gradient feedback.}
\label{fig:optimizer_prompt}
\end{figure}

\end{document}